%% file: 0main_cvpr.tex
\documentclass[final]{cvpr}

\usepackage{times}
\usepackage{epsfig}
\usepackage{graphicx}
\usepackage{amsmath}
\usepackage{amssymb}

\usepackage{bm}
\usepackage{url}
\usepackage{bbm}

\usepackage{color}
\usepackage{comment}
\usepackage{booktabs}
\usepackage{makecell}
\usepackage{multirow}
\usepackage{paralist}

\usepackage{makecell} %

\usepackage[ruled,vlined,linesnumbered]{algorithm2e}

\input{preambular2}

\usepackage[pagebackref=true,breaklinks=true,colorlinks,bookmarks=false]{hyperref}

\def\cvprPaperID{4378} %
\def\confYear{CVPR 2021}

\begin{document}

\title{General Multi-label Image Classification with Transformers}

\author{Jack Lanchantin, Tianlu Wang, Vicente Ordonez, Yanjun Qi\\
University of Virginia\\
{\tt\small \{jjl5sw,tianlu,vicente,yq2h\}@virginia.edu}
}

\maketitle
\input{0abstract}

\input{1intro}

\input{2back}

\input{3-1method}

\input{5-tables}

\input{5-1exp_standard}

\input{5-2exp_partial}

\input{5-3exp_extra}

\input{5-4partialAblation}

\input{4connecting}

\input{7conclusion}

\paragraph{Acknowledgements}
\input{A-funding}

{\small
\bibliographystyle{ieee_fullname}
\bibliography{cogat}
}

\input{9-1-Appendix}

\end{document}

%% file: preambular2.tex
\def\y{{\mathbf y}}

\newcommand{\modelname}{C-Tran}

\graphicspath{{..//}{./}}

\usepackage{placeins}

\usepackage{lipsum}

%% file: 0abstract.tex
\vspace{-15pt}
\begin{abstract}
Multi-label image classification is the task of predicting a set of labels corresponding to objects, attributes or other entities present in an image. In this work we propose the Classification Transformer (\modelname{}), a general framework for multi-label image classification that leverages Transformers to exploit the complex dependencies among visual features and labels. Our approach consists of a Transformer encoder trained to predict a set of target labels given an input set of masked labels, and visual features from a convolutional neural network. A key ingredient of our method is a label mask training objective that uses a ternary encoding scheme to represent the state of the labels as {positive}, {negative}, or {unknown} during training.
Our model shows state-of-the-art performance on challenging datasets such as COCO and Visual Genome. Moreover, because our model explicitly represents the uncertainty of labels during training, it is more general by allowing us to produce improved results for images with partial or extra label annotations during inference. We demonstrate this additional capability in the COCO, Visual Genome, News-500, and CUB image datasets.  %
\end{abstract}
\vspace{-5pt}

%% file: 1intro.tex
\section{Introduction}
\label{sec:intro}

Images in real-world applications generally portray many objects and complex situations. Multi-label image classification is a visual recognition task that aims to predict a set of labels corresponding to objects, attributes, or actions given an input image~\cite{elisseeff2002kernel, tsoumakas2006multi,wang2016cnn,wang2017multi,chen2017order,lee2018multi,ML-GCN}. %
This task goes beyond the more well studied single-label multi-class classification problem where the objective is to extract and associate image features with a single concept per image. In the multi-label setting, the output set of labels has some structure that reflects the structure of the world. For example, \textit{dolphin} is unlikely to co-occur with \textit{grass}, while \textit{knife} is more likely to appear next to a \textit{fork}. Effective models for multi-label classification aim to extract good visual features that are predictive of image labels, but also exploit the complex relations and dependencies between visual features and labels, and among labels themselves.

\begin{figure}[t]
\centering
\includegraphics[width=1.0\linewidth]{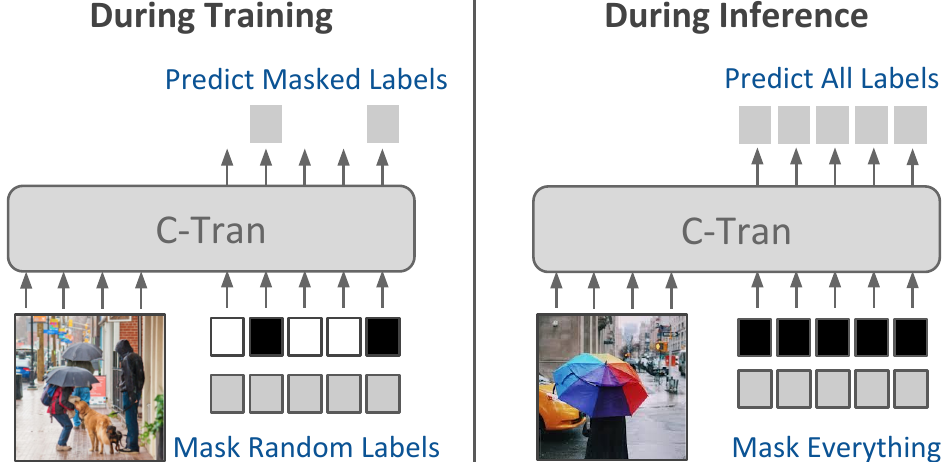}
\caption{We propose a transformer-based model for multi-label image classification that exploits dependencies among a target set of labels using an encoder transformer. During training, the model learns to reconstruct a partial set of labels given randomly masked input label embeddings and image features. During inference, our model can be conditioned only on visual input or a combination of visual input and partial labels, leading to superior results.}
\label{fig:lead}
\vspace{-5pt}
\end{figure}

\begin{figure*}[t]
\centering
\includegraphics[width=.98\textwidth]{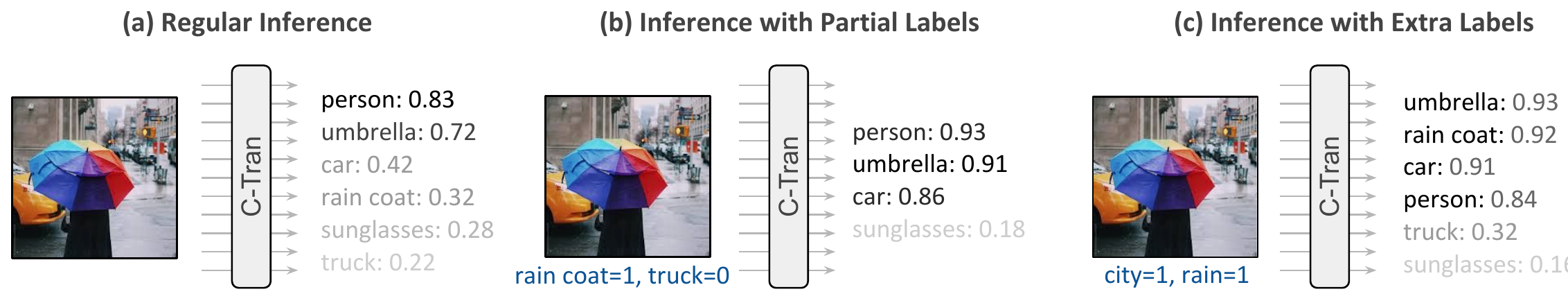}
\caption{
Different inference settings for general multi-label image classification: 
(a) Standard multi-label classification takes only image features as input. All labels are unknown $\y{}_{u}$.; 
(b) Classification under partial labels takes as input image features as well as a subset of the target labels that are known. The labels \textit{rain coat} and \textit{truck} are known labels $\y{}_{k}$, and all others are unknown labels $\y{}_{u}$;
(c) Classification under extra labels takes as input image features and some related extra information. The labels \textit{city} and \textit{rain} are known extra labels $\y{}^{e}_{k}$, and all others are unknown target labels $\y{}^{t}_{u}$.
} 
\label{fig:teaser} 
\vspace{-5pt}
\end{figure*}

To this end, we present the Classification Transformer (\modelname{}), a multi-label classification framework that leverages a Transformer encoder~\cite{vaswani2017attention}. Transformers have demonstrated a remarkable capability of being able to exploit complex and rich dependencies among sets of inputs using multiple layers of multi-headed self-attention operations. In our approach, a Transformer encoder is trained to reconstruct a set of target labels given an input set of masked label embeddings and a set of features obtained from a convolutional neural network. Unlike the Transformer encoders used for language modeling \cite{devlin2018bert}, \modelname{} uses a label mask training objective that allows us to represent the state of the labels as \emph{positive}, \emph{negative}, or \emph{unknown}. At test time, \modelname{} is able to predict a set of target labels using only input visual features by masking all the input labels as \emph{unknown}. Figure~\ref{fig:lead} gives an overview of this strategy. We demonstrate that this approach leads to superior results on a number of benchmarks compared to other recent approaches that exploit label relations using graph convolutional networks and other recently proposed strategies.

Beyond obtaining state-of-the-art results on the introduced regular multi-label classification task, we also claim that \modelname{} is a more general model for reasoning under prior label observations. Because our approach explicitly models the uncertainty of the labels during training, it can also be used at test time with partial or extra label annotations by setting the state of some of the labels as either \emph{positive} or \emph{negative} instead of masking them out as \emph{unknown}. For instance, consider the example shown in Figure~\ref{fig:teaser}(a) where a model is able to predict \emph{person} and \emph{umbrella} with relatively high accuracies, but is not confident for categories such as \emph{rain coat}, or \emph{car} that are clearly present in the picture. Suppose we know some labels and set them to their true positive (for \emph{rain coat}) or true negative (for \emph{truck}) values. Provided with this new information, the model is able to predict \emph{car} with a high confidence as it moves mass probability from \emph{truck} to \emph{car}, and predicts other objects such as \emph{umbrella} with even higher confidence than in the original predictions (Figure~\ref{fig:teaser}(b)). 
In general, we consider this setting as realistic since many images also have metadata in the form of extra labels such as location or weather information (Figure~\ref{fig:teaser}(c)).  This type of conditional inference is a much less studied problem. Our general approach to multi-label image classification with Transformers is able to naturally handle all these scenarios under a unified framework. We compare our results with a competing method relying on iterative inference~\cite{feedbackprop_CVPR_2018}, and against sensitive baselines, demonstrating superior results under variable amounts of partial or extra labels.

The benefits of our proposed framework can be summarized as follows: 
\begin{compactitem}
    \item Flexibility: It is the first model that can be deployed in multi-label image classification under arbitrary amounts of extra or partial labels. We use a unified model architecture and training method that lets users to apply our model easily in any setting.
    \item Accuracy: We evaluate our model on six datasets across three inference settings and achieve state-of-the-art results on all six.
    The label mask training strategy enhances the correlations between visual concepts leading to more accurate predictions.
    \item Interactivity: The use of state embeddings enables users to easily interact with the model and test any counterfactuals. \modelname{} can take human interventions as partial evidence and provides more interpretable and accurate predictions. 
\end{compactitem}

\begin{figure*}[t!]
\centering
\includegraphics[width=0.82\textwidth]{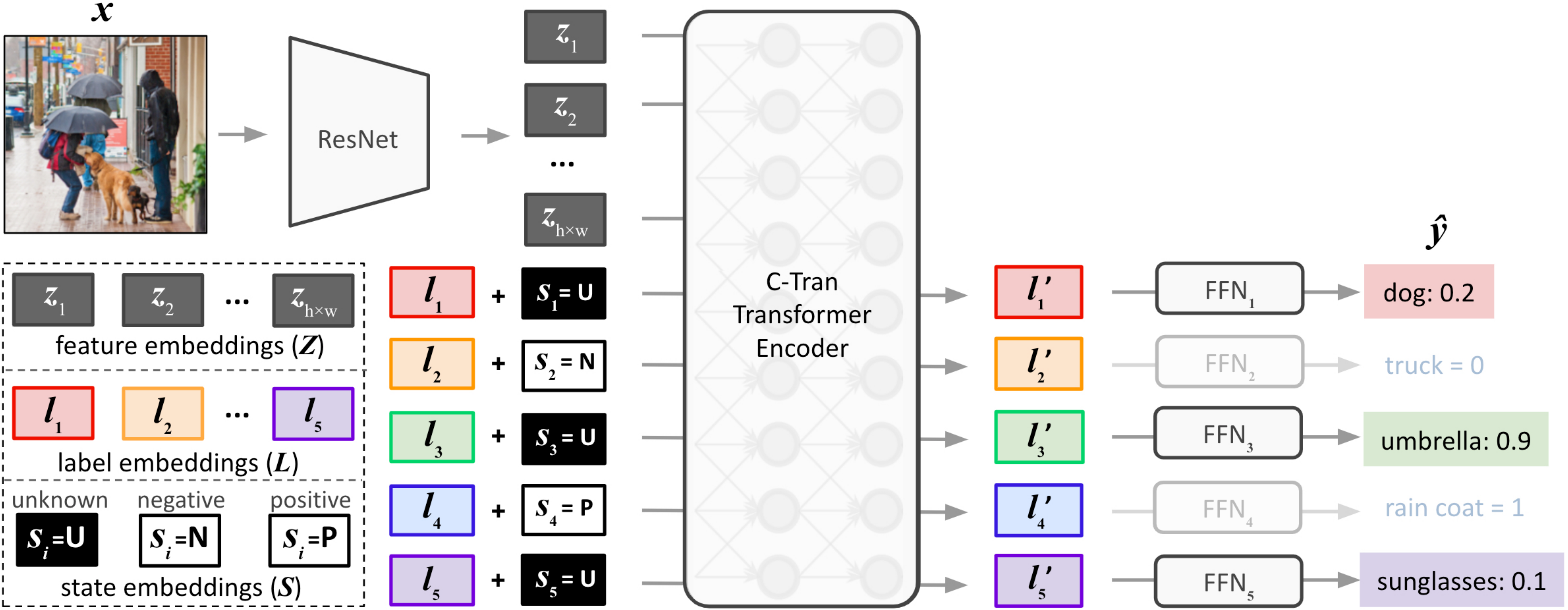}

\vspace{2pt}
\caption{
\small 
\modelname{} architecture and illustration of label mask training for general multi-label image classification. In this training image, the labels \textit{person}, \textit{umbrella}, and \textit{sunglasses} were randomly masked out and used as the unknown labels, $\y{}_u$. The labels \textit{rain coat} and \textit{truck} are used as the known labels, $\y{}_k$. Each unknown label is added the unknown state embedding U, and each known label is added its corresponding state embedding: negative (N) , or positive (P). The loss function is only computed on the unknown label predictions $\mathbf{\hat{y}}_u$.
}
\label{fig:framework}
\vspace{-5pt}
\end{figure*}

%% file: 2back.tex
\section{Problem Setup}
\label{sec:background}
In this section, we formally explain the three different multi-label image classification inference settings that we use to demonstrate the utility of our approach. 

\vspace{3pt}
\noindent \textbf{Regular Multi-label Classification.} 
In regular multi-label image classification, the goal is to predict a set of labels for an input image. Let $\mathbf{x}$ be an image, and $\mathbf{y}$ be a ground truth set of $\ell$ binary labels $\{{y}_1,{y}_2,...,{y}_{\ell}\}, y_i \in \{0,1\}$. The goal of multi-label classification is to construct a classifier, $f$, to predict a set of labels given an image so that: $\hat{\mathbf{y}} = f(\mathbf{x})$. 

\vspace{3pt}
\noindent \textbf{Inference with Partial Labels.}
While regular classification methods aim to predict the full set of $\ell$ labels given only an input image, some subset of labels $\mathbf{y}_{k} \subseteq \mathbf{y}$ may be observed, or known, at test time. This is also known as having partial labels available. For example, many images on the web are accompanied by some labeled text such as captions on social media. In this reformulated setting, the goal is to predict the unknown labels ($\mathbf{y}_{u} = \mathbf{y} \setminus \mathbf{y}_{k} $) given both the image \textit{and} the known labels during inference: $\hat{\mathbf{y}}_{u} = f(\mathbf{x},\mathbf{y}_{k})$. Note that we assume that all labels are properly annotated during training. This setting is specifically for \textit{inference} with partially annotated labels, and it differs from other works that tackle the problem of training models from partially annotated data \cite{xie2018partial,durand2019learning,kundu2020exploiting}. %

\vspace{3pt}
\noindent \textbf{Inference with Extra Labels.}
Similar to partially labeled images, there are many cases where we observe extra labels that describe the image, but are not part of the target label set. For example, we may know that an image was taken in a city. While ``city'' might not be one of the labels we want to predict, it can still alter our perception about what might be in the image. In this setting, we append any potential extra labels, denoted $\mathbf{y}^{e}$, to the target label set $\mathbf{y}^{t}$. If there are $\ell^{t}$ target labels, and $\ell^{e}$ potential extra labels, we now have a set of $\ell^{t}+\ell^{e}$ total labels that we train the model to predict. $\mathbf{y}$ now represents the concatenation of all target and extra labels. During inference, the known labels, $\mathbf{y}^{e}_{k}$, come from the set of extra labels, but we are only interested in evaluating the unknown target labels $\mathbf{y}^{t}_{u}$. In other words, during inference, we want to compute the following: $\hat{\mathbf{y}}^{t}_{u} = f(\mathbf{x},\mathbf{y}^{e}_{k})$. Again, we assume that all training images are fully annotated with their correct target and extra labels.

%% file: 3-1method.tex
\section{Method: \modelname{}}
\label{sec:method}
Considering the three inference settings described, we propose Classification Transformers (\modelname{}), a general multi-label classification framework that works in all three. During inference, our method predicts a set of unknown labels $\y{}_{u}$ given an input image $\mathbf{x}$ and a set of known labels $\y{}_{k}$. In regular inference no labels are known, in partial label inference some labels are known, and in extra label inference some labels external to the target set are known. In Sections \ref{sec:feat_lab_representation}-\ref{sec:classifier}, we introduce the \modelname{} architecture, and in Section \ref{sec:training_method}, we explain the label mask training procedure.

\subsection{Feature, Label, and State Embeddings}
\label{sec:feat_lab_representation}

\vspace{3pt}
\noindent \textbf{Image Feature Embeddings $\bm{Z}$:} Given input image $\mathbf{x} \in \mathbb{R}^{H\times W\times 3}$, the feature extractor outputs a tensor $\bm{Z} \in \mathbb{R}^{h\times w\times d}$, where $h,w,\textrm{ and }d$ are the output height, width, and channel, respectively. We can then consider each vector $\bm{z}_i \in \mathbb{R}^{d}$ from $\bm{Z}$, with $i$ ranging from $1$ to $P$ (where $P=h\times w$), to be representative of a subregion that maps back to patches in the original image space. 

\vspace{3pt}
\noindent \textbf{Label Embeddings $\bm{L}$:} For every image, we retrieve a set of label embeddings $\bm{L}$ = $\{\bm{l}_1,\bm{l}_2,...,\bm{l}_{\ell}\}$,  $\bm{l}_i \in \mathbb{R}^{d}$, which are representative of the $\ell$ possible labels in $\y{}$. Label embeddings are learned from an embedding layer of size $d \times \ell$.

\vspace{3pt}
\noindent \textbf{Adding Label Knowledge via State Embeddings $\bm{S}$:} 
In traditional architectures, there is no way to encode partially known or extra labels as input to the model. To address this drawback, we propose a technique to easily incorporate such information. Given label embedding $\bm{l}_i$, we simply add a ``state'' embedding vector, $\bm{s}_i \in \mathbb{R}^{d}$:
\begin{equation}
    \bm{\tilde{l}}_i = \bm{l}_i + \bm{s}_i,
    \label{eq:state_emb}
\end{equation}
where the $\bm{s}_i$ takes on one of three possible states: unknown (U), negative (N), or positive (P). For instance, if label $y_i$ is a known positive value prior to inference (meaning that we have prior knowledge that the label is present in the image), $\bm{s}_i$ is the positive embedding, P. The state embeddings are retrieved from a learned embedding layer of size $d \times 3$, where the unknown state vector (U) is fixed with all zeros.

State embeddings enable a user to (1) not use any prior information by adding the unknown embedding, (2), use partially labeled or extra information by adding the negative and positive embeddings to those labels, and (3) easily test interventions in the model by asking ``how does the prediction change if set this label to positive (negative)?''. We note that using prior information is completely optional as input to our model during testing, enabling it to also flexibly handle the regular inference setting.

\subsection{Modeling Feature and Label Interactions with a Transformer Encoder}
\label{sec:feat_lab_interactions}

To model the complex interactions between the image feature and embeddings, we develop our model based on a Transformer \cite{vaswani2017attention}. Transformers have proven to be a powerful mechanism for capturing rich dependency information between variables. 
Our formulation lets us to easily input the image feature and label embeddings jointly into a Transformer encoder. Transformer encoders are suitable because they are order invariant, allowing for any type of dependencies between all features and labels to be learned. 

Let $H=\{\bm{z}_{1},...,\bm{z}_{h \times w},\bm{\tilde{l}}_{1},...,\bm{\tilde{l}}_{\ell}\}$ be the set of embeddings that are input to the Transformer encoder. In Transformers, the importance, or weight, of embedding $\bm{h}_{j} \in H$ with respect to $\bm{h}_{i} \in H$ is learned through ``self-attention''. The attention weight, $\alpha^t_{ij}$ between embedding $i$ and $j$ is computed in the following manner. 
First, we compute a normalized scalar attention coefficient $\alpha_{ij}$ between embeddings $i$ and $j$. After computing the $\alpha_{ij}$ value for all $i$ and $j$ pairs, we update each $\bm{h}_{i}$ to $\bm{h}'_{i}$ using a weighted sum of all embeddings followed by a nonlinear ReLU layer:
\begin{gather}
    \alpha_{ij} =
    \textrm{softmax}\big((\mathbf{W}^q\bm{h}_i)^{\top} (\mathbf{W}^k\bm{h}_j)/\sqrt{\smash[b]d}\big) \\
    \bm{\bar{h}}_{i}= \sum_{j=1}^{M} \alpha_{i j} \mathbf{W}^v \bm{h}_{j} \\
    \bm{h}'_{i} =  \textrm{ReLU}(\bm{\bar{h}}_{i}\mathbf{W}^r + \bm{b}_1)\mathbf{W}^o + \bm{b}_2.
    \label{eq:ffn}
\end{gather}
where $\mathbf{W}^k$ is the key weight matrix, $\mathbf{W}^q$ is the query weight matrix,$\mathbf{W}^v$ is the value weight matrix, $\mathbf{W}^r$ and $\mathbf{W}^o$ are transformation matrices, and $\bm{b}_1$ and $\bm{b}_2$ are bias vectors.
This update procedure can be repeated for $L$ layers where the updated embeddings $\bm{h}'_{i}$ are fed as input to the successive Transformer encoder layer. The learned weight matrices $\{\mathbf{W}^k,\mathbf{W}^q,\mathbf{W}^v,\mathbf{W}^r,\mathbf{W}^o\} \in \mathbb{R}^{d\times d}$ are not shared between layers. We denote the final output of the Transformer encoder after $L$ layers as $H'=\{\bm{z}'_{1},...,\bm{z}'_{h \times w},\bm{l}'_{1},...,\bm{l}'_{\ell}\}$.

\subsection{Label Inference Classifier}
\label{sec:classifier}

Lastly, after feature and label dependencies are modeled via the Transformer encoder, a classifier makes the final label predictions. We use an independent feedforward network (FFN$_i$) for final label embedding $\bm{l}'_{i}$. FFN$_i$ contains a single linear layer, where weight $\mathbf{w}^c_i$ for label $i$ is a $1 \times d$ vector, and $\sigma$ is a simoid function:
\begin{equation}
    \hat{y}_i = \textrm{FFN}_i(\bm{l}'_{i}) = \sigma \big( (\mathbf{w}^c_i\cdot \bm{l}'_{i}) + b_i \big)
\end{equation}

\subsection{Label Mask Training (LMT)}
\label{sec:training_method}

State embeddings (Eq.~\ref{eq:state_emb}) lets us easily incorporate known labels as input to \modelname{}.  However, we want our model to be flexible enough to handle any amount of known labels during inference. To solve this problem, we introduce a novel training procedure called Label Mask Training (LMT) that forces the model to learn label correlations, and allows \modelname{} to generalize to any inference setting.

Inspired by the Cloze task \cite{taylor1953cloze} and the BERT ``masked language model'' \cite{devlin2018bert} which learn semantic information by predicting missing words from their context, we implement a similar procedure. During training, we randomly \textit{mask} a certain amount of labels, and use the ground truth of the other labels (via state embeddings) to predict the masked labels. This differs from masked language model training in that we have a fixed set of inputs (all possible labels) and we randomly mask a subset of them for each sample.

Given that there are $\ell$ possible labels, the number of ``unknown'' (i.e. masked) labels for a particular sample, $n$, is chosen at random between $0.25\ell$ and $\ell$. Then, $n$ unknown labels, denoted $\y{}_{u}$, are sampled randomly from all possible labels $\y{}$. The unknown state embedding is added to each unknown label. The rest are ``known'' labels, denoted $\y{}_{k}$ and the corresponding ground truth state embedding (positive or negative) is added to each. We call these known labels because the ground truth value is used as input to \modelname{} alongside the image. Our model predicts the unknown labels $\y{}_{u}$, and binary cross entropy is used to update the model parameters. 

By masking random amounts of unknown labels (and therefore using random amounts of known labels) during training, the model learns many possible known label combinations. This allows the \modelname{} to be used in any inference setting where there may be arbitrary amounts of known information. 

We mask out at least 0.25$\ell$ labels for each training samples for several reasons. First, most masked language model training methods mask out around 15\% of the words \cite{devlin2018bert,brown2020language}. Second, we want our model to be able to incorporate anywhere from 0 to 0.75$\ell$ known labels during inference. We assume that knowing more than 75\% of the labels is an unrealistic inference scenario.

Essentially, our label mask training pipeline tries to minimize the following loss approximately:
\begin{equation}
L=\sum_{n=1}^{N_{tr}} \mathbb{E}_{p(\y_k)} \{ \text{CE}( \hat{\y}_u^{(n)},\y_u^{(n)}) | \y_k \},
\label{eq:loss}
\end{equation}
where CE represents the cross entropy loss function. $\mathbb{E}_{p(\y_k)} (\cdot | \y_k )$ denotes to calculate the expectation regarding the probability distribution of known labels:  $\y_k$.

\subsection{Implementation Details}
\label{model_details}
\noindent \textbf{Image Feature Extractor.\,} For fair comparisons, we use the same image size and pretrained feature extractor as the previous state-of-the-art in each setting. For all datasets except CUB, we use the ResNet-101 \cite{he2016deep} pretrained on ImageNet \cite{imagenet_cvpr09} as the feature extractor (for CUB, we use the same as \cite{koh2020concept}). Since the output dimension of ResNet-101 is $2048$, we set our embedding size $d$ as $2048$. Following  \cite{chen2019learning,chen2020knowledge}, training are images resized to $640 \times 640$ and randomly cropped to $576 \times 576$ with random horizontal flips. Testing images are center cropped instead. The output of the ResNet-101 model is an $18$$\times$$18$$\times$$d$ tensor, so there are a total of 324 feature embedding vectors, $\bm{z}_i \in \mathbb{R}^{d}$.

\vspace{3pt}
\noindent \textbf{Transformer Encoder.\,}
In order to allow a particular embedding to attend to multiple other embeddings (or multiple groups), \modelname{} uses 4 attention heads \cite{vaswani2017attention}. 
We use a $L$=3 layer Transformer with a residual layer \cite{he2016deep} around each embedding update and layer norm \cite{ba2016layer}.

\vspace{3pt}
\noindent \textbf{Optimization.\,}
For training, Adam \cite{kingma2014adam} is used as the optimizer with betas=$(0.9, 0.999)$ and weight decay=$0$. We train the models with a batch size of $16$ and a learning rate of $10^{-5}$.  We use dropout \cite{goodfellow2016deep} of $p=0.1$ for regularization.

%% file: 5-tables.tex
\begin{table*}[ht]
\begin{center}
\small
\begin{tabular}{|l||l|l|l|l|l|l|l||l|l|l|l|l|l|}
\hline
    \multirow{2}{*}{} &
  \multicolumn{6}{c}{All} &
  \multicolumn{1}{c||}{} &
  \multicolumn{5}{c}{Top 3} &
  \multicolumn{1}{c|}{} 
  \\ \cline{2-14}\cline{2-14}

& mAP  & CP   & CR   & CF1  & OP   & OR   & OF1  & CP   & CR   & CF1  & OP   & OR   & OF1  \\ \hline\hline
CNN-RNN \cite{wang2016cnn}          & 61.2 & -    & -    & -    & -    & -    & -    & 66.0 & 55.6 & 60.4 & 69.2 & 66.4 & 67.8 \\
RNN-Attention \cite{wang2017multi}  & -    & -    & -    & -    & -    & -    & -    & 79.1 & 58.7 & 67.4 & 84.0 & 63.0 & 72.0 \\
Order-Free RNN \cite{chen2017order} & -    & -    & -    & -    & -    & -    & -    & 79.1 & 58.7 & 67.4 & 84.0 & 63.0 & 72.0 \\
ML-ZSL \cite{lee2018multi}          & -    & -    & -    & -    & -    & -    & -    & 74.1 & 64.5 & 69.0 & -    & -    & -    \\
SRN \cite{SRN}                      & 77.1 & 81.6 & 65.4 & 71.2 & 82.7 & 69.9 & 75.8 & 85.2 & 58.8 & 67.4 & 87.4 & 62.5 & 72.9 \\
ResNet101 \cite{he2016deep}         & 77.3 & 80.2 & 66.7 & 72.8 & 83.9 & 70.8 & 76.8 & 84.1 & 59.4 & 69.7 & 89.1 & 62.8 & 73.6 \\
Multi-Evidence \cite{ge2018multi}   & -    & 80.4 & 70.2 & 74.9 & 85.2 & 72.5 & 78.4 & 84.5 & 62.2 & 70.6 & 89.1 & 64.3 & 74.7 \\
ML-GCN \cite{ML-GCN}                & 83.0 & 85.1 & 72.0 & 78.0 & 85.8 & 75.4 & 80.3 & 89.2 & 64.1 & 74.6 & 90.5 & 66.5 & 76.7 \\
SSGRL \cite{chen2019learning}       & 83.8 & \textbf{89.9} & 68.5 & 76.8 & \textbf{91.3} & 70.8 & 79.7 & \textbf{91.9} & 62.5 & 72.7 & \textbf{93.8} & 64.1 & 76.2 \\
KGGR \cite{chen2020knowledge}       & 84.3 & 85.6 & 72.7 & 78.6 & 87.1 & 75.6 & 80.9 & 89.4 & 64.6 & 75.0 & 91.3 & 66.6 & 77.0 \\
\hline\hline
\modelname{}&
  \textbf{85.1} & 86.3 & \textbf{74.3} & \textbf{79.9} &  87.7 & \textbf{76.5} & \textbf{81.7} & 90.1 & \textbf{65.7} & \textbf{76.0} & 92.1 & \textbf{71.4} &  \textbf{77.6} \\\hline
\end{tabular}
\end{center}
\caption{Results of \textit{regular inference} on COCO-80 dataset. The threshold is set to $0.5$ to compute precision, recall and F1 scores (\%). Our method consistently outperforms previous methods across multiple metrics under the settings of all and top-3 predicted labels. Best results are shown in bold. ``-'' denotes that the metric was not reported.  \label{tab:mlc_standard}}
\label{tab:mlc_standard_coco80}
\end{table*}

\begin{table*}[ht]
\begin{center}
\small
\begin{tabular}{|l||l|l|l|l|l|l|l||l|l|l|l|l|l|}
\hline
\multirow{2}{*}{} &
  \multicolumn{6}{c}{All} &
  \multicolumn{1}{c||}{} &
  \multicolumn{5}{c}{Top 3} &
  \multicolumn{1}{c|}{} 
  \\ \cline{2-14}
  
   & mAP  & CP   & CR   & CF1  & OP            & OR   & OF1  & CP            & CR   & CF1  & OP   & OR   & OF1  \\ \hline\hline

ResNet101\cite{he2016deep} \hspace{16pt}  & 30.9 & 39.1 & 25.6 & 31.0 & 61.4          & 35.9 & 45.4 & 39.2          & 11.7 & 18.0 & 75.1 & 16.3 & 26.8 \\
ML-GCN \cite{ML-GCN}          & 32.6 & 42.8 & 20.2 & 27.5 & 66.9          & 31.5 & 42.8 & 39.4          & 10.6 & 16.8 & 77.1 & 16.4 & 27.1 \\
SSGRL \cite{chen2019learning} & 36.6 &  -   &  -   &-     & -             & -    & -    &   -           &   -  &     -&  -   & -    &-    \\
KGGR \cite{chen2020knowledge} & 37.4 & 47.4 & 24.7 & 32.5 & \textbf{66.9}          & 36.5 & 47.2 & 48.7          & 12.1 & 19.4 & 78.6 & 17.1 & 28.1 \\
\hline\hline
\modelname{} &
  \textbf{38.4} &
  \textbf{49.8} &
  \textbf{27.2} &
  \textbf{35.2} &
  \textbf{66.9} &
  \textbf{39.2} &
  \textbf{49.5} &
  \textbf{51.1} &
  \textbf{12.5} &
  \textbf{20.1} &
  \textbf{80.2} &
  \textbf{17.5} &
  \textbf{28.7} \\\hline
\end{tabular}
\end{center}
\caption{Results of \textit{regular inference} on VG-500 dataset. All metrics and setups are the same as~Table~\ref{tab:mlc_standard_coco80}. Our method achieves notable improvement over previous methods.
\label{tab:mlc_standard_vg500}}
\end{table*}

\begin{table*}[htb]
\begin{center}
\small
\setlength\tabcolsep{3pt}
\begin{tabular}{|l||llll||llll||llll||llll|}
\hline
 & \multicolumn{4}{c||}{COCO-80} & \multicolumn{4}{c||}{VG-500} & \multicolumn{4}{c||}{NEWS-500} & \multicolumn{4}{c|}{COCO-1000} \\ 
 Partial Labels Known ($\epsilon$) \hspace{2pt}
& 0\% & 25\% & 50\% & 75\% & 0\% & 25\% & 50\% & 75\% & 0\% & 25\% & 50\% & 75\% & 0\% & 25\% & 50\% & 75\% \\ \hline\hline
Feedbackprop \cite{feedbackprop_CVPR_2018} & 80.1 & 80.6 & 80.8 & 80.9 & 29.6 & 30.1 & 30.8 & 31.6 & 14.7 & 21.1 & 23.7 & 25.9 & 29.2 & 30.1 & 31.5 & 33.0 \\ 
\modelname{} & \textbf{85.1} & \textbf{85.2} & \textbf{85.6} & \textbf{86.0} & \textbf{38.4} & \textbf{39.3} & \textbf{40.4} & \textbf{41.5} & \textbf{18.1} & \textbf{29.7} & \textbf{35.5} & \textbf{39.4} & \textbf{34.3} & \textbf{35.9} & \textbf{37.4} & \textbf{39.1} \\ \hline
\end{tabular}
\end{center}
\caption{Results of \textit{inference with partial labels} on four multi-label image classification datasets. Mean average precision score (\%) is reported. Across four simulated settings where different amounts of partial labels are available ($\epsilon$), our method significantly outperforms the competing method. With more partial labels available, we achieve larger improvement.}
\label{tab:mlc_partial}
\end{table*}

\begin{table}[htb]
\begin{center}
\small
\setlength\tabcolsep{2.5pt} 
\begin{tabular}{|l||cccc|}
\hline
Extra Label Groups Known ($\epsilon$) & 0\%              & 36\%            & 54\%            & 71\%            \\ \hline\hline
Standard \cite{koh2020concept}            & 82.7          & 82.7          & 82.7          & 82.7             \\
Multi-task \cite{koh2020concept}          & \textbf{83.8} & 83.8          & 83.8          & 83.8         \\
ConceptBottleneck \cite{koh2020concept}  & 80.1          & 87.0          & 93.0          & 97.5         \\
\hline\hline
\modelname{}   & \textbf{83.8} & \textbf{90.0} & \textbf{97.0} & \textbf{98.0} \\ \hline
\end{tabular}
\end{center}
\caption{Results of \textit{inference with extra labels} on CUB-312 dataset. We report the accuracy score (\%) for the 200 multi-class target labels. We achieve similar or greater accuracy than the baselines across all amounts of known extra label groups.}
\label{tab:mlc_auxiliary}
\vspace{-5pt}
\end{table}

%% file: 5-1exp_standard.tex
\section{Experimental Setup and Results}
\label{sec:partial_mlc_result}
In the following subsections, we explain the datasets, baselines, and results for the three multi-label classification inference settings.

\subsection{Regular Inference}
\noindent \textbf{Datasets.}
We use two large-scale regular multi-label classification datasets: COCO-80 and VG-500. COCO \cite{Lin2014MicrosoftCC}, is a commonly used large scale dataset for multi-label classification, segmentation, and captioning. It contains $122,218$ images containing common objects in their natural context. The standard multi-label formulation for COCO, which we call COCO-80, includes $80$ object class annotations for each image. We use $82,081$ images as training data and evaluate all methods on a test set consisting of $40,137$ images.
The Visual Genome dataset \cite{Krishna2016VisualGC},  contains $108,077$ images with object annotations covering thousands of categories. Since the label distribution is very sparse, we only consider the $500$ most frequent objects and use the VG-500 subset introduced in \cite{chen2020knowledge}. VG-500 consists of $98,249$ training images and $10,000$ test images.

\vspace{3pt}
\noindent \textbf{Baselines and Metrics.}
For COCO-80, we compare to ten well known multi-label classification methods. For VG-500 we compare to four previous methods that used this dataset.

Referencing previous works \cite{ML-GCN,chen2019learning,chen2020knowledge}, we employ several metrics to evaluate the proposed method and existing methods. Concretely, we report the average per-class precision (CP), recall (CR), F1 (CF1) and the average overall precision (OP), recall (OR), F1 (OF1), under the setting that a predicted label is positive if the output probability is greater than $0.5$. We also report the mean average precision (mAP). A detailed explanation of the metrics are shown in the Appendix. For fair comparisons to previous works \cite{ge2018multi, SRN}, we also consider the setting where we evaluate the Top-3 predicted labels following. In general, \textbf{mAP}, \textbf{OF1}, and \textbf{CF1} are the most important metrics \cite{ML-GCN}.

\vspace{3pt}
\noindent \textbf{Results.}
\modelname{} achieves state-of-the-art performance across almost all metrics on both datasets, as shown in Table~\ref{tab:mlc_standard_coco80} and Table~\ref{tab:mlc_standard_vg500}. Considering that COCO-80 and VG-500 are two widely studied multi-label datasets, absolute mAP increases of 0.8 and 1.0, respectively, can be considered notable improvements. Importantly, we do not use any predefined feature and label relationship information (e.g. pretrained word embeddings). This signals that our method can effectively lean the relationships.

%% file: 5-2exp_partial.tex
\subsection{Inference with Partial Labels }

\noindent \textbf{Datasets.}
We use four datasets to validate our approach in the partial label setting. In all four datasets, we simulate four amounts of partial labels during inference. More specifically, for each testing image, we select $\epsilon$ percent of labels as known. $\epsilon$ is set to 0\% / 25\% / 50\% / 75\% in our experiments. $\epsilon$=0\% denotes no known labels, and is equivalent to the regular inference setting.

In addition to COCO-80 and VG-500, we benchmark our method on two more multi-label image classification datasets. Wang et al. \cite{feedbackprop_CVPR_2018} derived the top $1000$ frequent words from the accompanying captions of COCO images to use as target labels, which we call COCO-1000. There are $82,081$ images for training, and $5,000$ images for validation and testing, respectively. We expect that COCO-1000 provides more and stronger dependencies compared to COCO-80. We also use the NEWS-500 dataset \cite{feedbackprop_CVPR_2018}, which was collected from the BBC News. Similar to COCO-1000, the target label set consists of $500$ most frequent nouns derived from image captions. There are $151,873$ images for training, $10,304$ for validation and $10,451$ for testing. 

\vspace{3pt}
\noindent \textbf{Baselines and Metrics.}
Feedback-prop \cite{feedbackprop_CVPR_2018} is an inference method introduced for partial label inference that make use of arbitrary amount of known labels. This method backpropagates the loss on the known labels to update the intermediate image representations during inference. We use the LF method on ResNet-101 Convolutional Layer 13 from \cite{feedbackprop_CVPR_2018}.  
We compute the mean average precision (mAP) score of predictions on unknown labels. 

\vspace{3pt}
\noindent \textbf{Results.}
As shown in Table~\ref{tab:mlc_partial}, \modelname{} outperforms Feedbackprop, in all $\epsilon$ percentages of partially known labels on all datasets. In addition, as the percentage of partial labels increases, the improvement of \modelname{} over Feedbackprop also increases. 
These results demonstrate that our method can effectively leverage known labels and is very flexible with the amount of known labels.
Feedbackprop updates image features which implicitly encode some notion of label correlation. \modelname{}, instead, explicitly models the correlations of between labels and features, leading to improved results especially when partial labels are known. On the other hand, Feedback-prop requires careful hyperparameter tuning on a separate validation set and needs time-consuming iterative feature updates. Our method does not require any hyerparameter tuning and just needs a standard one-pass inference. We include qualitative examples in Appendix, demonstrating that effectiveness of our method.

%% file: 5-3exp_extra.tex
\subsection{Inference with Extra Labels }

\noindent \textbf{Datasets.}
For the extra label setting, we use the Caltech-UCSD Birds-200-2011 (CUB) dataset \cite{welinder2010caltech}.  It contains 9,430 training samples and 2,358 testing samples. We conduct a multi-classification task with 200 bird species on this dataset. Multi-class classification is a specific instantiation of multi-label classification, where the target classes are mutually exclusive. In other words, each image has only one correct label. 
We use the processed CUB dataset from Koh et al. \cite{koh2020concept} where they include $112$ extra labels related to bird species. We call this dataset CUB-312.
They further cluster extra labels into 28 groups and use varying amounts of known groups at inference time.  To make a fair comparison, we consider four different amounts of extra label groups for inference: 0 group (0\%), 10 groups (36\%), 15 groups (54\%), and 20 groups (71\%).

\vspace{3pt}
\noindent \textbf{Baselines and Metrics.}
Concept Bottleneck Models \cite{koh2020concept} incorporate the extra labels as intermediate labels ( ``concepts'' in the original paper). They construct bottleneck layer to first predict the extra labels, and then use those predictions to predict bird species. 
I.e., if we let $\mathbf{y}^{e}$ be the extra information labels, \cite{koh2020concept} predicts the target class labels $\mathbf{y}^{t}$ using the following computation graph: $\mathbf{x} \rightarrow \mathbf{y}^{e} \rightarrow \mathbf{y}^{t}$. 
We also consider two baselines from \cite{koh2020concept}. The first is standard multi-layer perception model that does not use a bottleneck layer. The second is a multi-task learning model that predicts the target and concept labels jointly.
For fair comparison, we use the same feature extraction method for all methods, Inception-v3 \cite{szegedy2016rethinking}. Since the target task is multi-class, we evaluate the target predictions using accuracy scores. 

\vspace{3pt}
\noindent \textbf{Results.}
Table \ref{tab:mlc_auxiliary} shows that \modelname{} achieves an improved accuracy over Concept Bottleneck models on the CUB-312 task when using any amount of extra label groups. Notably, the multi-task learning model produces the best performing results when $\epsilon$=0. However, it is not able to incorporate known extra labels (i.e., $\epsilon>$0). \modelname{} instead, consistently achieves the best performance.
Additionally, we can test interventions, or counterfactuals, using \modelname{}. For example, ``grey beak'' is one of the extra labels, and we can set the state embedding of ``grey beak'' to be positive or negative and observe the change in bird class predictions.  We provide samples of extra label intervention in the Appendix.

%% file: 5-4partialAblation.tex
\subsection{Ablation and Model Analysis}
\label{sec:ablation}

In this section, we conduct ablation studies to analyze the contributions of each \modelname{} component. We examine two settings: regular inference (equivalent to 0\% known partial labels) and 50\% known partial label inference. We evaluate on four datasets: COCO-80, VG-500, NEWS-500, and COCO-1000. First, we remove the image features $\mathbf{Z}$ and predict unknown labels given only known labels. This experiment, \modelname{} (no image), tells us how much information model can learn just from labels. Table~\ref{tab:ablation_component} shows that we get relatively high mean average precision scores on some datasets (NEWS-500 and COCO-1000). This indicates that even without image features, \modelname{} is able to effectively learn rich dependencies from label annotations .

\begin{figure*}[ht]
\centering
\fbox{
\begin{minipage}[b]{0.47\textwidth}
    \includegraphics[width=1\linewidth]{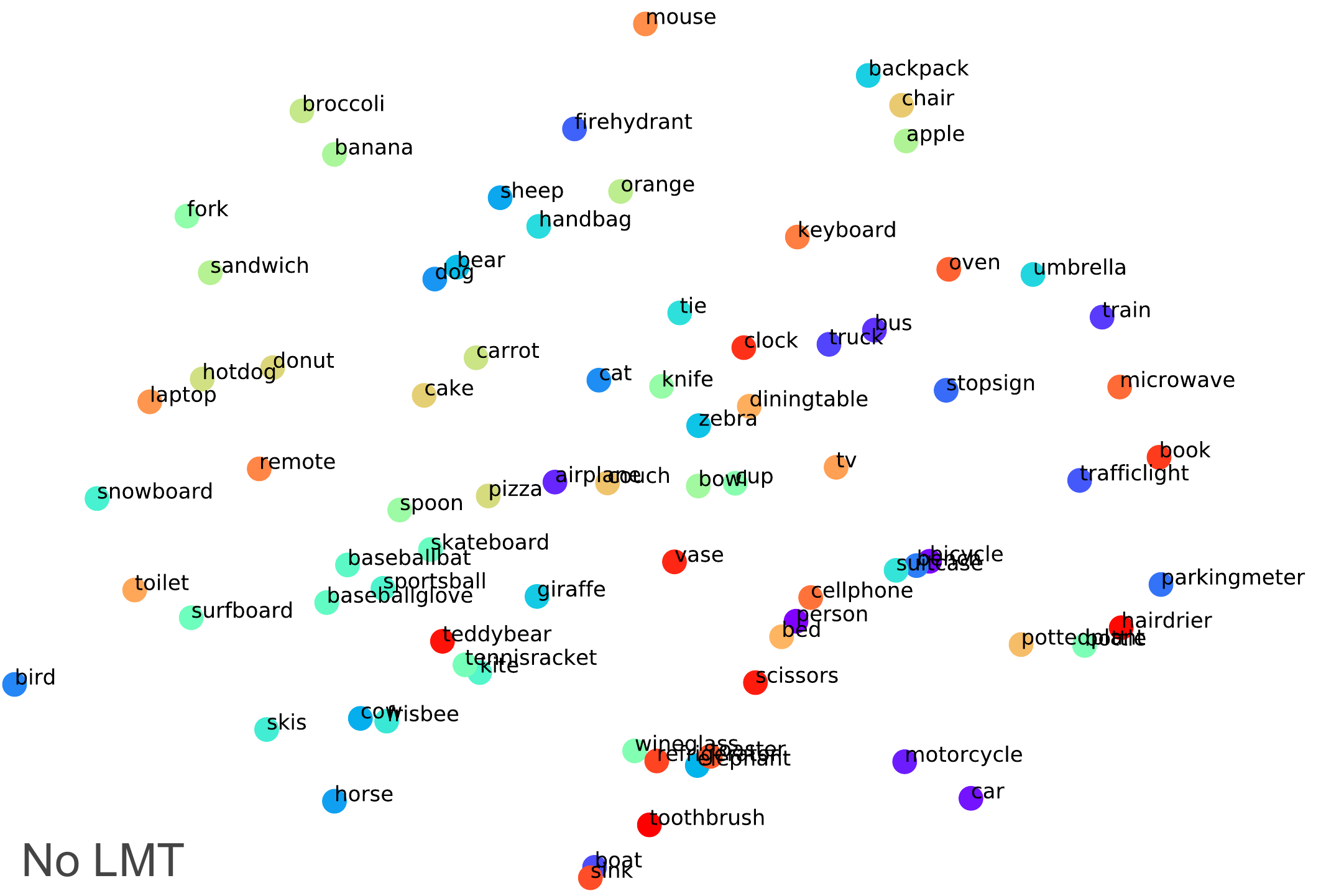}
\end{minipage}
}
\hspace{5pt}
\fbox{
\begin{minipage}[b]{0.45\textwidth}
\includegraphics[width=1\linewidth]{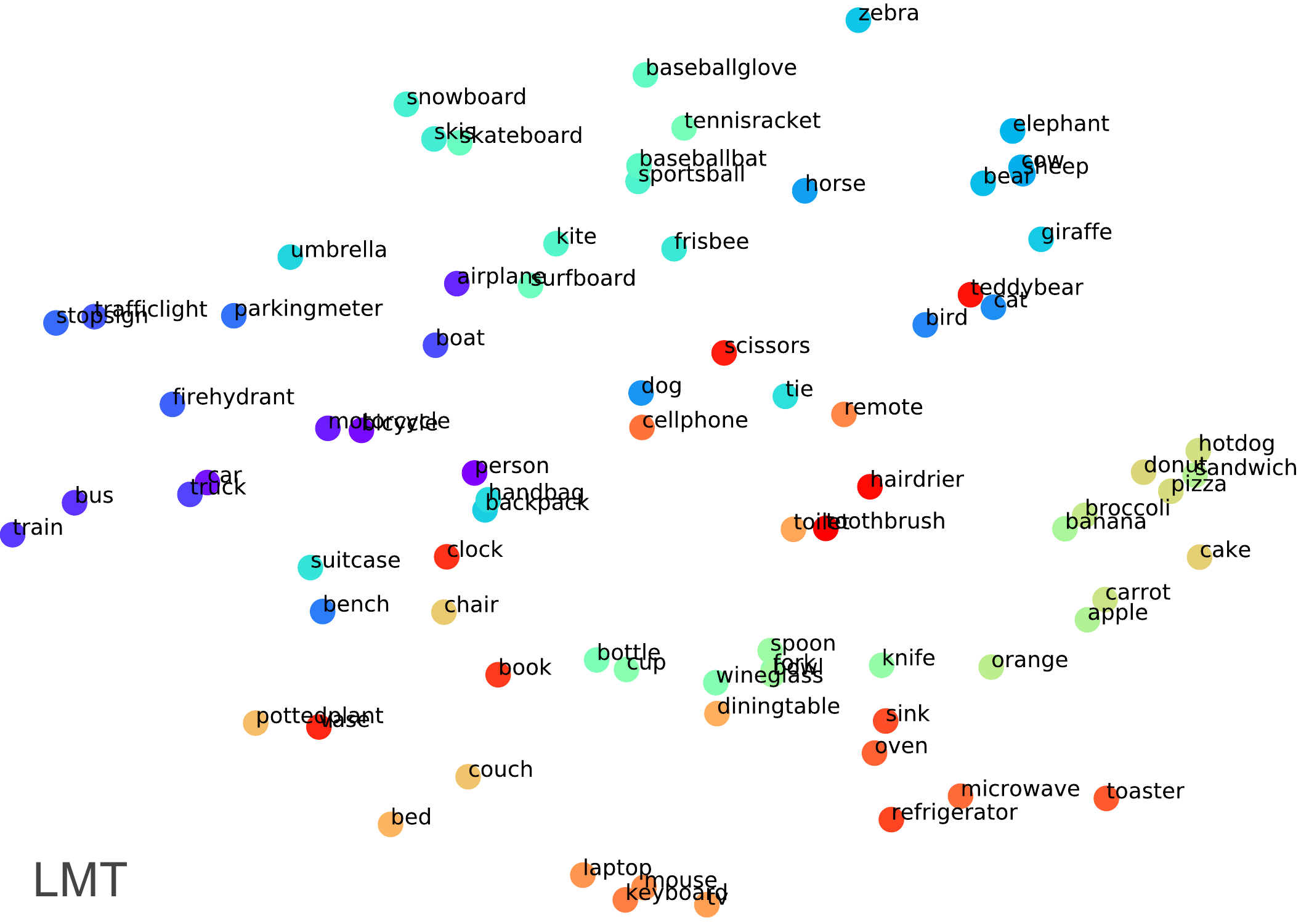}
\end{minipage}
}
\vspace{5pt}
\small \caption{Comparison of the learned label embeddings for COCO-80 using t-SNE. The left figure shows the embedding projections without using label mask training (LMT), and the right shows with LMT.  Labels are colored using the COCO object categorization. We can see that using label mask training produces much semantically stronger label representations.}
\label{fig:tsne}
\vspace{-2pt}
\end{figure*}

Second, we remove the label mask training procedure to test the effectiveness of this technique. More specifically, we remove all label state embeddings, $\mathbf{S}$; thus all labels are unknown during training. Table~\ref{tab:ablation_component} shows that for both settings, regular (0\%) and 50\% partial labels known, the performance drops without label mask training. This signifies two critical findings of label mask training: (1) it helps with dependency learning as we see improvement when no partial labels are available during inference. This is particularly true for datasets that have strong label co-occurrences, such as NEWS-500 and COCO-1000. (2) given partial labels, it can significantly improve prediction accuracy. We provide a t-SNE plot \cite{maaten2008visualizing} of the label embeddings learned with or without label mask training. As revealed in Figure~\ref{fig:tsne}, embeddings learned with label mask training show more meaningful semantic topology; objects belonging to the same group are clustered together.

We also analyze the importance of the number of Transformer layers, $L$, in the regular inference setting for COCO-80. Mean average precision scores for $2$, $3$, and $4$ layers were $85.0$, $85.1$, and $84.3$, respectively. This indicates : (1) our method is fairly robust to the number of Transformer layers, (2) multi-label classification likely doesn't require a very large number of layers like some other NLP tasks, which use 96 layers \cite{brown2020language}. 
While we show \modelname{} is a powerful method in many multi-label classification settings, we recognize that Transformer layers are memory-intensive for a large number of inputs. This limits the number of possible labels $\ell$ in our model. Using four NVIDIA Titan X GPUs, the upper bound of $\ell$ is around 2000 labels. However, it is possible to increase the number of labels. We currently use the ResNet-101 output channel size ($d=2048$) for our Transformer hidden layer size. This can be linearly mapped to a smaller number. Additionally, we could apply one of the Transformer variations that have been proposed to model very large input sizes \cite{dai2019transformer,sukhbaatar2019adaptive}.

\vspace{10pt}
\begin{table}[t]
\begin{center}
\small
\setlength\tabcolsep{1.5pt}
\begin{tabular}{|c|cc|cc|cc|cc|}
\hline
\multirow{2}{*}{\makecell{Partial Labels \\
Known ($\epsilon$) }} & \multicolumn{2}{l|}{\footnotesize COCO-80} & \multicolumn{2}{l|}{\footnotesize VG-500} & \multicolumn{2}{l|}{\footnotesize NEWS-500} & \multicolumn{2}{l|}{\footnotesize COCO-1000} \\
& \footnotesize 0\%  & \footnotesize 50\% & \footnotesize 0\%  & \footnotesize 50\% & \footnotesize 0\%  & \footnotesize 50\% & \footnotesize 0\%  & \footnotesize 50\% \\ \hline\hline
C-Tran (no image)    & \footnotesize 3.60  & \footnotesize 21.7 & \footnotesize 2.70 & \footnotesize 24.6  & \footnotesize 6.50  & \footnotesize 33.3 & \footnotesize 1.50  & \footnotesize 27.8 \\
C-Tran (no LMT)      & \footnotesize 84.8 & \footnotesize 85.0 & \footnotesize 38.3 & \footnotesize 38.8 & \footnotesize 16.9 & \footnotesize 17.1 & \footnotesize 33.1 & 3\footnotesize 4.0 \\
C-Tran & \footnotesize \textbf{85.1}     & \footnotesize \textbf{85.6}     & \footnotesize \textbf{38.4}     & \footnotesize \textbf{40.4}    & \footnotesize \textbf{18.1}      & \footnotesize \textbf{35.5}     & \footnotesize \textbf{34.3}      & \footnotesize \textbf{37.4}      \\ \hline
\end{tabular}
\end{center}
\caption{\modelname{} component ablation results. Mean average precision score (\%) is reported. 
Our proposed Label Mask Training technique (LMT) improves the performance, especially when partial labels are available.}
\label{tab:ablation_component}
\end{table}

%% file: 4connecting.tex
\vspace{-10pt}
\section{Related Work}
\noindent \textbf{Multi-label Image Classification.}
Multi-label classification (MLC) is gaining popularity due to its relevance in real world applications. Recently, \cite{stock2018convnets} showed that the remaining error in ImageNet is not due to the feature extraction, but rather that ImageNet is annotated with single labels even when some images depict more than one object.

Recent literature addressing multi-label classification roughly fall into four groups.  
\textit{(1) Conditional Prediction:} The first type, autoregressive models \cite{dembczynski2010bayes,read2009classifier,wang2016cnn,nam2017maximizing} estimate the true joint probability of output labels given the input by using the chain rule, predicting one label at a time.  
\textit{(2) Shared Embedding Space:} The second group learns to project input features and output labels into a shared latent embedding space \cite{yeh2017learning,bhatia2015sparse}. 
\textit{(3) Structured Output:} The third kind describes label dependencies using structured output inference formulation  \cite{lafferty2001conditional,tsochantaridis2005large,SPEN,guo2011multi,Liq,10.5555/3020751.3020796,nimmagadda2015multi}. 
\textit{(4) Label Graph Formulation:} Several recent studies \cite{ML-GCN,lanchantin2019neural,chen2019learning,chen2020knowledge} used graph neural networks to model label dependency an obtained state-of-the-art results.  All methods relied on knowledge-based graphs being built from label co-occurrence statistics. 
Our proposed model is most similar to \textit{(4)}, but it does not need extra knowledge to build a graph and can automatically learn the label dependency.

\vspace{3pt}
\noindent \textbf{Inference with Partial Labels}
Wang et al. proposed feedbackprop, a new inference strategy to handle any set of partial labels at test time~\cite{feedbackprop_CVPR_2018}. The core idea is to optimize intermediate image representations according to \textit{known} labels and then predict the \textit{unknown} labels based on the updated representations. However, this requires many iterations at inference time, resulting in a significantly slower classifier. Additionally, the model is never exposed to partial evidence during training, which limits the potential improvement. Several methods \cite{koperski2020plugin,hu2016learning} utilize partial labels using a fixed set of labels. However, these cannot generalize to arbitrary sets of known labels. In more realistic inference settings, there may be any subset of known labels available during inference. If there are $\ell$ total labels, then the number of known labels, $n$=$ |\mathbf{y}_{k}|$ ranges from 0 to $\ell$-$1$. The number of possible known label sets is then $\binom{\ell}{n}$.
\modelname{}, instead integrates a novel representation indicating each label state as \textit{positive}, \textit{negative} or \textit{unknown}. This representation enables us to leverage partial signals into the model training, and make our model compatible with any known label set during inference. Notably, \modelname{} is the first learning method that can exploit arbitrary amounts of partial evidence during both training and inference.

Many works tackle the problem of partial label multi-label classification \textit{training} \cite{xie2018partial,durand2019learning,kundu2020exploiting}. While this sounds similar to our setting, there are several key distinctions. First, these methods focus on the case of ``partial annotations'', which means that they assume not all labels are annotated correctly during training. We assume that all labels are correctly annotated during training.
Second, partial label training methods cannot be easily extended to the partial label inference setting. In other words, these methods can just take images as input and fail to incorporate extra useful information (partial/extra labels) during inference.

\vspace{3pt}
\noindent \textbf{Inference with Extra Labels}
\cite{koh2020concept} introduces Concept Bottleneck Models which incorporates intermediate concept labels as a bottleneck layer for the target label classification. 
Similar to \cite{koperski2020plugin}, this model assumes that the concept labels are a fixed set. While interpretability is an advantage, bottleneck models \cite{lampert2009learning,kumar2009attribute} rely on the assumption that the manually curated concepts are sufficient features for target class prediction, contradicting the feature learning approach of deep learning.
\modelname{}, uses state embeddings instead of a concept bottleneck layer to represent each concept as \textit{known} (positive or negative) or \textit{unknown}. This representation enables \modelname{} to leverage partial labels (concepts) during training, and make our model compatible with any known labels (concepts) during inference. Importantly, we do not have any assumptions of the size of labels (concepts) to be known during inference.

%% file: 7conclusion.tex
\vspace{-5pt}
\section{Conclusion}
\label{sec:conclude}

This paper proposes a novel deep learning method, called \modelname{}, for a wide variety of ``multi-label image classification'' applications. Our approach is easy to implement, requires no extra resources, and can effectively leverage any amount of partial or extra labels during inference.  \modelname{} learns sample-adaptive interactions through attention and can discover how the labels attend to different parts of an input image. We showcase the effectiveness of our approach in regular multi-label classification settings and multi-label classification with partially observed or extra labels. \modelname{} outperforms all state-of-the-art methods in all scenarios. We further provide a quantitative and qualitative analysis showing that \modelname{} boosts the performance by explicitly modeling the interactions between target labels and between image features and target labels. As the next steps, we plan to extend \modelname{} to hierarchical scene categorization applications. We also plan to explore the design of better training strategies to make \modelname{} generalize to settings where some labels have never been observed in training.

%% file: A-funding.tex
This work was partly supported by the National Science Foundation under NSF CAREER award No. 1453580 to Y.Q. and a Leidos gift award to V.O. Any opinions, findings and conclusions or recommendations expressed in this material are those of the author(s) and do not necessarily reflect those of the National Science Foundation.

%% file: 9-1-Appendix.tex
\appendix 
\section{Appendix}

\subsection{Qualitative Examples}

\begin{figure*}
\centering
\includegraphics[width=.9\textwidth]{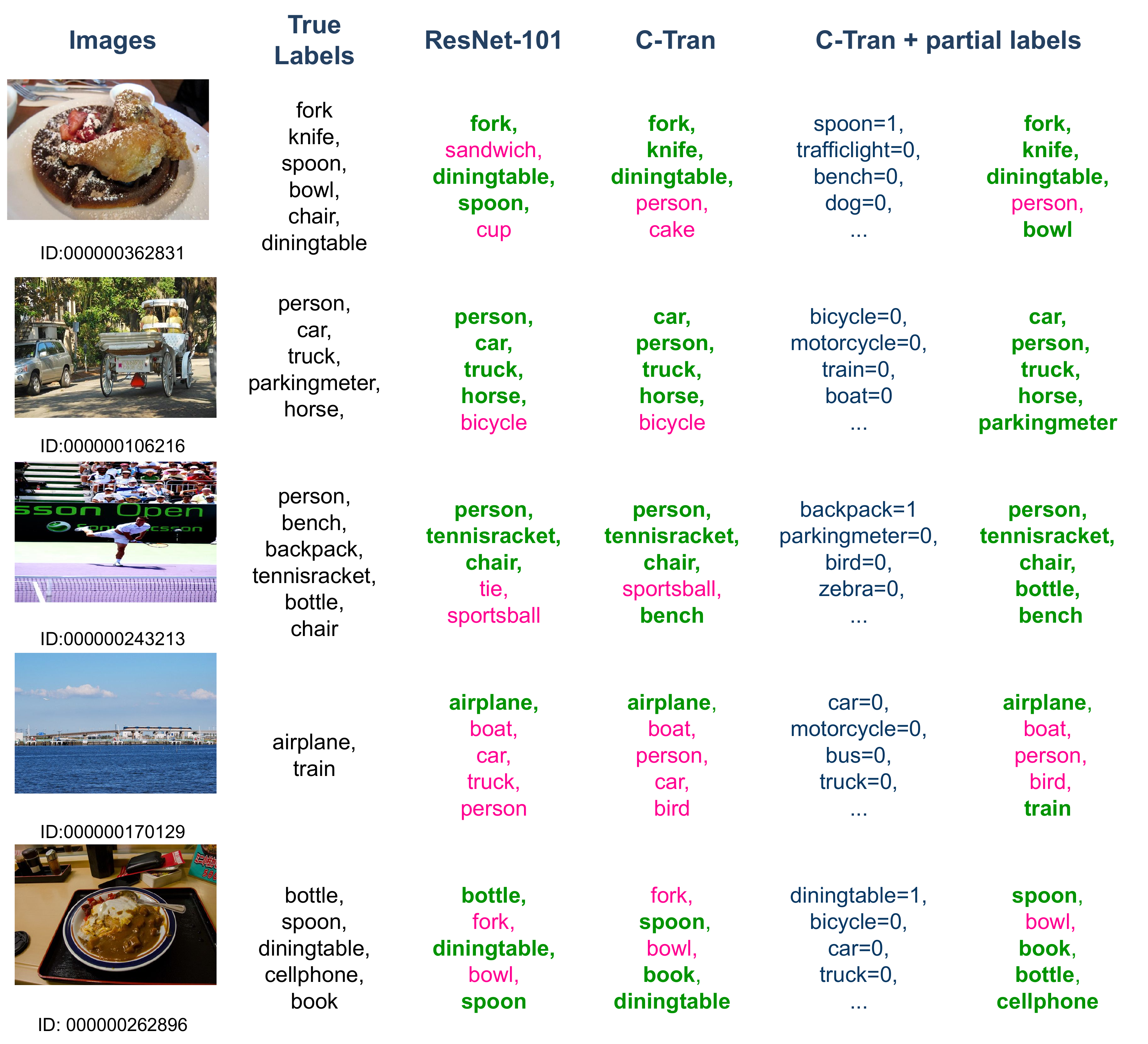}
\caption{Qualitative examples of \modelname{} + partial labels on the COCO-80 dataset. In the last column, we use $\epsilon=25\%$ partial labels, some of which are shown. Correctly predicted labels are in bold.}
\label{fig:coco_samples}
\end{figure*}

\textbf{Inference with Partial Labels\,\,}
In Figure~\ref{fig:coco_samples}, we show qualitative results on COCO-80 demonstrating the use of partial labels. In these examples, we first show the predictions for ResNet-101, as well as \modelname{} without using partial labels. The last column shows the \modelname{} predictions when using $\epsilon=25\%$ partial labels (which is 21 labels for COCO-80) as observed, or known prior to inference. For many examples, certain labels cannot be predicted well without using partial labels.

\begin{figure*}
\centering
\includegraphics[width=.9\textwidth]{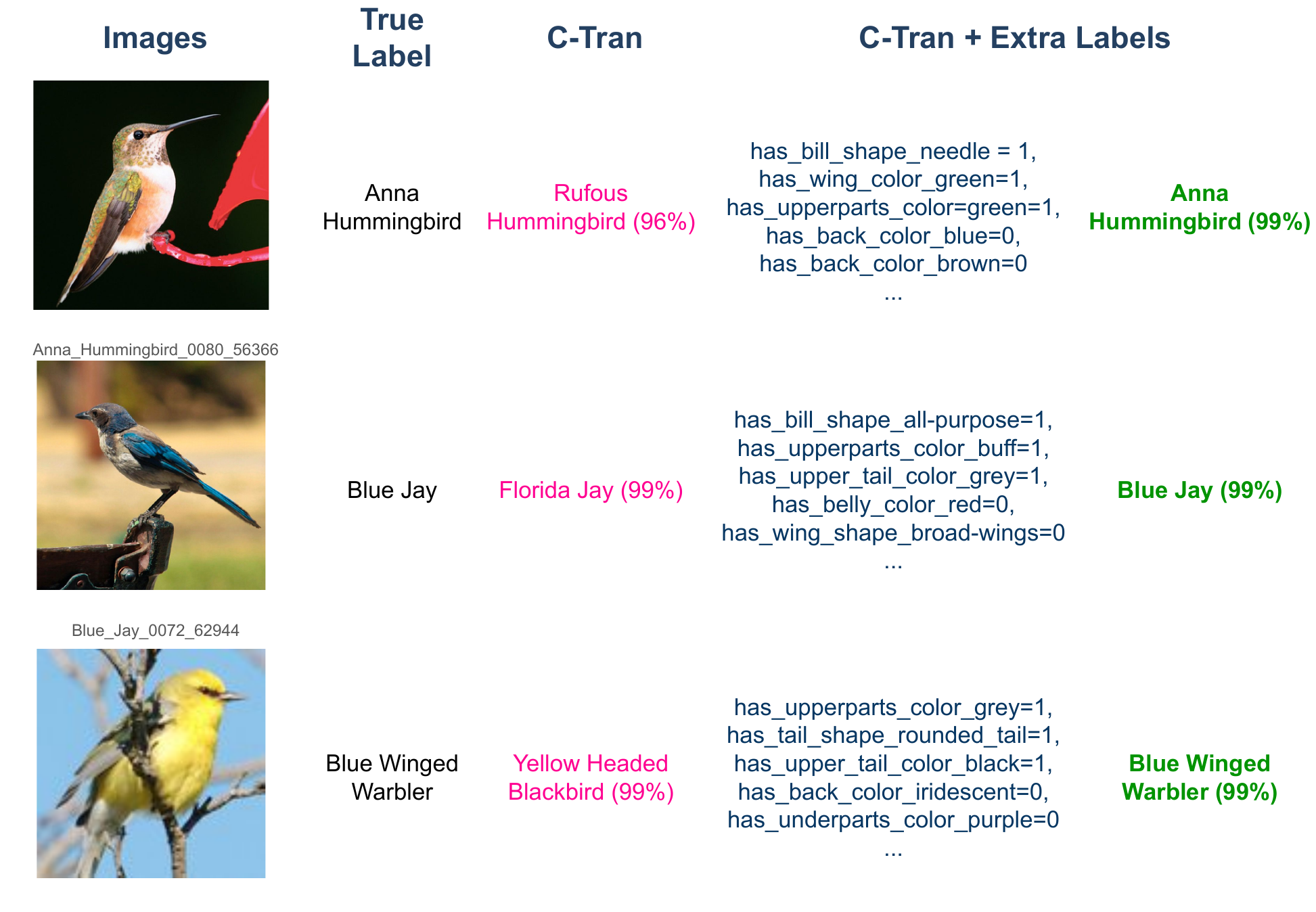}
\caption{Qualitative examples of \modelname{} + extra labels on the CUB-312 dataset. In the last column, we use $\epsilon=54\%$ extra labels, some of which are shown.}
\label{fig:cub_samples}
\vspace{-50pt}
\end{figure*}

\textbf{Inference with Extra Labels\,\,}
In Figure~\ref{fig:cub_samples}, we show qualitative results on CUB-312 demonstrating the use of extra labels. In the CUB-312 dataset, the extra labels are high level concepts of bird species that are not target labels. In these examples, we first show the predictions for \modelname{} without using extra labels labels, and the last column shows the \modelname{} predictions when using $\epsilon=54\%$ of the extra labels (which is 60 labels for CUB-312) as observed, or known prior to inference. We can see that many bird species predictions are completely changed after using the extra labels as input to our model.

\subsection{Detailed Diagram of \modelname{} Settings\,\,}

\begin{figure*}[t]
\centering
\includegraphics[width=1.0\linewidth]{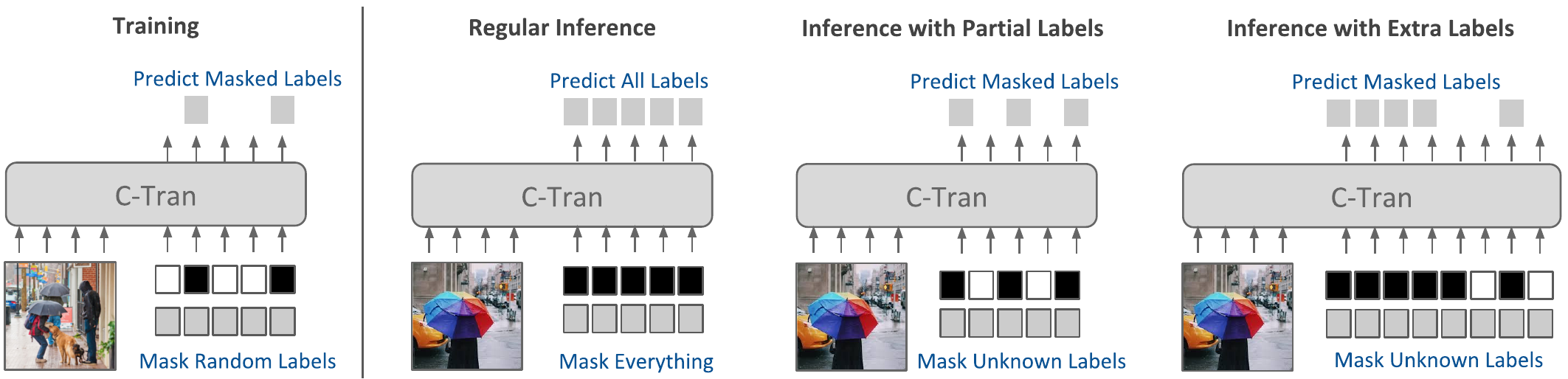}
\caption{Detailed example of the general training method and three different inference settings where \modelname{} can be applied.
}
\label{fig:teaser_full}
\vspace{-5pt}
\end{figure*}

In Figure~\ref{fig:teaser_full} shows a detailed diagram of all possible training and inference settings used in our paper, and how \modelname{} is used in each setting. By using the same random mask training, we can apply our model to any of the three inference settings.

\subsection{Multi-Label Classification Metrics}

\begin{equation}
\begin{aligned} \mathrm{OP} &=\frac{\sum_{i} N_{i}^{c}}{\sum_{i} N_{i}^{p}} \\ \mathrm{OR} &=\frac{\sum_{i} N_{i}^{c}}{\sum_{i} N_{i}^{g}} \\ \mathrm{OF} 1 &=\frac{2 \times \mathrm{OP} \times \mathrm{OR}}{\mathrm{OP}+\mathrm{OR}} \end{aligned}
\end{equation}

\begin{equation}
\begin{aligned} \mathrm{CP} &=\frac{1}{C} \sum_{i} \frac{N_{i}^{c}}{N_{i}^{p}} \\ \mathrm{CR} &=\frac{1}{C} \sum_{i} \frac{N_{i}^{c}}{N_{i}^{g}} \\ \mathrm{CF} 1 &=\frac{2 \times \mathrm{CP} \times \mathrm{CR}}{\mathrm{CP}+\mathrm{CR}} \end{aligned}
\end{equation}

where $C$ is the number of labels, $N_{i}^{c}$ is true positives for the $i$-th label, $N_{i}^{p}$ is the total number of images for which the $i$-th label is predicted, and $N_{i}^{g}$ is the number of ground truth images for the $i$-th label.

\subsection{More Discussions of \modelname{}\,\,}

\textbf{Connecting to Transformers and BERT\,\,}

Our proposed method, \modelname{}, draws much inspiration from works in natural language processing. The transformer model \cite{vaswani2017attention} proposed ``self attention'' for natural language translation. Self attention allows each word in the target sentence to attend to all other words (both in the source sentence and the target sentence) for translation. \cite{devlin2018bert} introduced BERT for language modeling. BERT uses self attention with masked words to pretrain a language model. 

Self attention and BERT are both examples of complete graphs, but on sentences rather than image features and labels.  \modelname{} uses the same self-attention mechanisms as \cite{vaswani2017attention} and \cite{devlin2018bert}, but instead of using only the word embeddings from a sentence, we use feature and label embeddings. 

In computer vision, \cite{carion2020end} used Transformers for object detection. Our method varies in several distinct ways. First, we are primarily interested in using partial evidence for image classification, and our unique state embeddings allow \modelname{} to use such evidence. Second, we model image and label features jointly in a Transformer encoder, whereas \cite{carion2020end} use an encoder/decoder framework. Our method allows the image features to be updated conditioned on the labels, which is a key characteristic of our model. 

\textbf{Connecting to Graph Based Neural Relational Learning}
Another line of recent works employ object localization techniques\cite{bbox,hcp} or attention mechanism\cite{Wang_2017_ICCV,SRN} to locate semantic meaningful regions and try to identify underlying relations between regions and outputs. However, these methods either require expensive bounding box annotations or merely get regions of interest roughly due to the lack of label supervision. One recent study by \cite{10.1109/ICCV.2011.6126300} also showed that modeling the associations between image feature regions and labels helps to improve multi-label performance. In our work, \modelname{} uses graph attentions and enables each target label to attend differentially to relevant parts of an input image.

For multi-label classfication(MLC), \cite{guo2011multi} formulate MLC using a label graph and they introduced a conditional dependency SVM where they first trained separate classifiers for each label given the input and all other true labels and used Gibbs sampling to find the optimal label set. The main drawback is that this method requires separate classifiers for each label.
\cite{su2013multilabel} proposes a method to label the pairwise edges of randomly generated label graphs, and requires some chosen aggregation method over all random graphs. The authors introduce the idea that variation in the graph structure shifts the inductive bias of the base learners. One recent study \cite{LAMP} used graph neural networks for multi-label classification on sequential inputs. The proposed method models the label-to-label dependencies using GNNs, however, does not represent input features and labels in one coherent graph. 
A key aspect of \modelname{} is that the Transformer encoder can be viewed as a fully connected graph which is able to learn any relationships between features and labels. The Transformer attention mechanism can be regarded as a form of graph ensemble learning \cite{hara2016analysis}.
Above all, previous methods using graphs to model label dependencies do not allow for partial evidence information to be included in the prediction.

%% file: 0main_cvpr.bbl
\begin{thebibliography}{10}\itemsep=-1pt

\bibitem{ba2016layer}
Jimmy~Lei Ba, Jamie~Ryan Kiros, and Geoffrey~E Hinton.
\newblock Layer normalization.
\newblock {\em arXiv preprint arXiv:1607.06450}, 2016.

\bibitem{SPEN}
David Belanger, Bishan Yang, and Andrew McCallum.
\newblock End-to-end learning for structured prediction energy networks.
\newblock In {\em Proceedings of the 34th International Conference on Machine
  Learning-Volume 70}, pages 429--439. JMLR. org, 2017.

\bibitem{bhatia2015sparse}
Kush Bhatia, Himanshu Jain, Purushottam Kar, Manik Varma, and Prateek Jain.
\newblock Sparse local embeddings for extreme multi-label classification.
\newblock In {\em Advances in neural information processing systems}, pages
  730--738, 2015.

\bibitem{brown2020language}
Tom~B Brown, Benjamin Mann, Nick Ryder, Melanie Subbiah, Jared Kaplan, Prafulla
  Dhariwal, Arvind Neelakantan, Pranav Shyam, Girish Sastry, Amanda Askell,
  et~al.
\newblock Language models are few-shot learners.
\newblock {\em arXiv preprint arXiv:2005.14165}, 2020.

\bibitem{carion2020end}
Nicolas Carion, Francisco Massa, Gabriel Synnaeve, Nicolas Usunier, Alexander
  Kirillov, and Sergey Zagoruyko.
\newblock End-to-end object detection with transformers.
\newblock {\em arXiv preprint arXiv:2005.12872}, 2020.

\bibitem{chen2017order}
Shang-Fu Chen, Yi-Chen Chen, Chih-Kuan Yeh, and Yu-Chiang~Frank Wang.
\newblock Order-free rnn with visual attention for multi-label classification.
\newblock {\em arXiv preprint arXiv:1707.05495}, 2017.

\bibitem{chen2020knowledge}
Tianshui Chen, Liang Lin, Xiaolu Hui, Riquan Chen, and Hefeng Wu.
\newblock Knowledge-guided multi-label few-shot learning for general image
  recognition.
\newblock {\em IEEE Transactions on Pattern Analysis and Machine Intelligence},
  2020.

\bibitem{chen2019learning}
Tianshui Chen, Muxin Xu, Xiaolu Hui, Hefeng Wu, and Liang Lin.
\newblock Learning semantic-specific graph representation for multi-label image
  recognition.
\newblock {\em arXiv preprint arXiv:1908.07325}, 2019.

\bibitem{ML-GCN}
Zhao-Min Chen, Xiu-Shen Wei, Peng Wang, and Yanwen Guo.
\newblock {Multi-Label Image Recognition with Graph Convolutional Networks}.
\newblock In {\em The IEEE Conference on Computer Vision and Pattern
  Recognition (CVPR)}, 2019.

\bibitem{dai2019transformer}
Zihang Dai, Zhilin Yang, Yiming Yang, Jaime Carbonell, Quoc~V Le, and Ruslan
  Salakhutdinov.
\newblock Transformer-xl: Attentive language models beyond a fixed-length
  context.
\newblock {\em arXiv preprint arXiv:1901.02860}, 2019.

\bibitem{dembczynski2010bayes}
Krzysztof Dembczynski, Weiwei Cheng, and Eyke H{\"u}llermeier.
\newblock Bayes optimal multilabel classification via probabilistic classifier
  chains.
\newblock In {\em .}, 2010.

\bibitem{imagenet_cvpr09}
J. Deng, W. Dong, R. Socher, L.-J. Li, K. Li, and L. Fei-Fei.
\newblock {ImageNet: A Large-Scale Hierarchical Image Database}.
\newblock In {\em CVPR09}, 2009.

\bibitem{devlin2018bert}
Jacob Devlin, Ming-Wei Chang, Kenton Lee, and Kristina Toutanova.
\newblock Bert: Pre-training of deep bidirectional transformers for language
  understanding.
\newblock {\em arXiv preprint arXiv:1810.04805}, 2018.

\bibitem{durand2019learning}
Thibaut Durand, Nazanin Mehrasa, and Greg Mori.
\newblock Learning a deep convnet for multi-label classification with partial
  labels.
\newblock In {\em Proceedings of the IEEE Conference on Computer Vision and
  Pattern Recognition}, pages 647--657, 2019.

\bibitem{elisseeff2002kernel}
Andr{\'e} Elisseeff and Jason Weston.
\newblock A kernel method for multi-labelled classification.
\newblock In {\em Advances in neural information processing systems}, pages
  681--687, 2002.

\bibitem{ge2018multi}
Weifeng Ge, Sibei Yang, and Yizhou Yu.
\newblock Multi-evidence filtering and fusion for multi-label classification,
  object detection and semantic segmentation based on weakly supervised
  learning.
\newblock In {\em Proceedings of the IEEE Conference on Computer Vision and
  Pattern Recognition}, pages 1277--1286, 2018.

\bibitem{goodfellow2016deep}
Ian Goodfellow, Yoshua Bengio, and Aaron Courville.
\newblock {\em Deep learning}.
\newblock MIT press, 2016.

\bibitem{guo2011multi}
Yuhong Guo and Suicheng Gu.
\newblock Multi-label classification using conditional dependency networks.
\newblock In {\em IJCAI Proceedings-International Joint Conference on
  Artificial Intelligence}, volume~22, page 1300, 2011.

\bibitem{hara2016analysis}
Kazuyuki Hara, Daisuke Saitoh, and Hayaru Shouno.
\newblock Analysis of dropout learning regarded as ensemble learning.
\newblock In {\em International Conference on Artificial Neural Networks}.
  Springer, 2016.

\bibitem{he2016deep}
Kaiming He, Xiangyu Zhang, Shaoqing Ren, and Jian Sun.
\newblock Deep residual learning for image recognition.
\newblock In {\em Proceedings of the IEEE conference on computer vision and
  pattern recognition}, pages 770--778, 2016.

\bibitem{hu2016learning}
Hexiang Hu, Guang-Tong Zhou, Zhiwei Deng, Zicheng Liao, and Greg Mori.
\newblock Learning structured inference neural networks with label relations.
\newblock In {\em Proceedings of the IEEE Conference on Computer Vision and
  Pattern Recognition}, pages 2960--2968, 2016.

\bibitem{kingma2014adam}
Diederik~P Kingma and Jimmy Ba.
\newblock Adam: A method for stochastic optimization.
\newblock {\em arXiv preprint arXiv:1412.6980}, 2014.

\bibitem{koh2020concept}
Pang~Wei Koh, Thao Nguyen, Yew~Siang Tang, Stephen Mussmann, Emma Pierson, Been
  Kim, and Percy Liang.
\newblock Concept bottleneck models.
\newblock {\em arXiv preprint arXiv:2007.04612}, 2020.

\bibitem{koperski2020plugin}
Michal Koperski, Tomasz Konopczynski, Rafal Nowak, Piotr Semberecki, and Tomasz
  Trzcinski.
\newblock Plugin networks for inference under partial evidence.
\newblock In {\em The IEEE Winter Conference on Applications of Computer
  Vision}, pages 2883--2891, 2020.

\bibitem{Krishna2016VisualGC}
Ranjay Krishna, Yuke Zhu, Oliver Groth, Justin Johnson, Kenji Hata, Joshua
  Kravitz, Stephanie Chen, Yannis Kalantidis, Li-Jia Li, David~A. Shamma,
  Michael~S. Bernstein, and Li Fei-Fei.
\newblock Visual genome: Connecting language and vision using crowdsourced
  dense image annotations.
\newblock {\em International Journal of Computer Vision}, 123:32--73, 2016.

\bibitem{kumar2009attribute}
Neeraj Kumar, Alexander~C Berg, Peter~N Belhumeur, and Shree~K Nayar.
\newblock Attribute and simile classifiers for face verification.
\newblock In {\em 2009 IEEE 12th international conference on computer vision},
  pages 365--372. IEEE.

\bibitem{kundu2020exploiting}
Kaustav Kundu and Joseph Tighe.
\newblock Exploiting weakly supervised visual patterns to learn from partial
  annotations.
\newblock {\em Advances in Neural Information Processing Systems}, 33, 2020.

\bibitem{lafferty2001conditional}
John Lafferty, Andrew McCallum, and Fernando~CN Pereira.
\newblock Conditional random fields: Probabilistic models for segmenting and
  labeling sequence data.
\newblock {\em .}, 2001.

\bibitem{lampert2009learning}
Christoph~H Lampert, Hannes Nickisch, and Stefan Harmeling.
\newblock Learning to detect unseen object classes by between-class attribute
  transfer.
\newblock In {\em 2009 IEEE Conference on Computer Vision and Pattern
  Recognition}, pages 951--958. IEEE, 2009.

\bibitem{lanchantin2019neural}
Jack Lanchantin, Arshdeep Sekhon, and Yanjun Qi.
\newblock Neural message passing for multi-label classification.
\newblock In {\em Joint European Conference on Machine Learning and Knowledge
  Discovery in Databases}, pages 138--163. Springer, 2019.

\bibitem{LAMP}
Jack Lanchantin, Arshdeep Sekhon, and Yanjun Qi.
\newblock Neural message passing for multi-label classification.
\newblock {\em ECML}, abs/1904.08049, 2019.

\bibitem{lee2018multi}
Chung-Wei Lee, Wei Fang, Chih-Kuan Yeh, and Yu-Chiang Frank~Wang.
\newblock Multi-label zero-shot learning with structured knowledge graphs.
\newblock In {\em Proceedings of the IEEE conference on computer vision and
  pattern recognition}, pages 1576--1585, 2018.

\bibitem{Liq}
Qiang Li, Maoying Qiao, Wei Bian, and Dacheng Tao.
\newblock Conditional graphical lasso for multi-label image classification.
\newblock In {\em CVPR}, pages 2977--2986, 06 2016.

\bibitem{10.5555/3020751.3020796}
Xin Li, Feipeng Zhao, and Yuhong Guo.
\newblock Multi-label image classification with a probabilistic label
  enhancement model.
\newblock In {\em Proceedings of the Thirtieth Conference on Uncertainty in
  Artificial Intelligence}, UAI'14, pages 430--439, Arlington, Virginia, USA,
  2014. AUAI Press.

\bibitem{Lin2014MicrosoftCC}
Tsung-Yi Lin, Michael Maire, Serge~J. Belongie, James Hays, Pietro Perona, Deva
  Ramanan, Piotr Doll{\'a}r, and C.~Lawrence Zitnick.
\newblock Microsoft coco: Common objects in context.
\newblock In {\em ECCV}, 2014.

\bibitem{maaten2008visualizing}
Laurens van~der Maaten and Geoffrey Hinton.
\newblock Visualizing data using t-sne.
\newblock {\em Journal of machine learning research}, 9(Nov):2579--2605, 2008.

\bibitem{nam2017maximizing}
Jinseok Nam, Eneldo~Loza Menc{\'\i}a, Hyunwoo~J Kim, and Johannes
  F{\"u}rnkranz.
\newblock Maximizing subset accuracy with recurrent neural networks in
  multi-label classification.
\newblock In {\em Advances in Neural Information Processing Systems}, pages
  5419--5429, 2017.

\bibitem{nimmagadda2015multi}
Tejaswi Nimmagadda and Anima Anandkumar.
\newblock Multi-object classification and unsupervised scene understanding
  using deep learning features and latent tree probabilistic models.
\newblock {\em arXiv preprint arXiv:1505.00308}, 2015.

\bibitem{read2009classifier}
Jesse Read, Bernhard Pfahringer, Geoff Holmes, and Eibe Frank.
\newblock Classifier chains for multi-label classification.
\newblock {\em Machine Learning and Knowledge Discovery in Databases}, pages
  254--269, 2009.

\bibitem{stock2018convnets}
Pierre Stock and Moustapha Cisse.
\newblock Convnets and imagenet beyond accuracy: Understanding mistakes and
  uncovering biases.
\newblock In {\em Proceedings of the European Conference on Computer Vision
  (ECCV)}, pages 498--512, 2018.

\bibitem{su2013multilabel}
Hongyu Su and Juho Rousu.
\newblock Multilabel classification through random graph ensembles.
\newblock In {\em Asian Conference on Machine Learning}, pages 404--418, 2013.

\bibitem{sukhbaatar2019adaptive}
Sainbayar Sukhbaatar, Edouard Grave, Piotr Bojanowski, and Armand Joulin.
\newblock Adaptive attention span in transformers.
\newblock {\em arXiv preprint arXiv:1905.07799}, 2019.

\bibitem{szegedy2016rethinking}
Christian Szegedy, Vincent Vanhoucke, Sergey Ioffe, Jon Shlens, and Zbigniew
  Wojna.
\newblock Rethinking the inception architecture for computer vision.
\newblock In {\em Proceedings of the IEEE conference on computer vision and
  pattern recognition}, pages 2818--2826, 2016.

\bibitem{taylor1953cloze}
Wilson~L Taylor.
\newblock “cloze procedure”: A new tool for measuring readability.
\newblock {\em Journalism quarterly}, 30(4):415--433, 1953.

\bibitem{tsochantaridis2005large}
Ioannis Tsochantaridis, Thorsten Joachims, Thomas Hofmann, and Yasemin Altun.
\newblock Large margin methods for structured and interdependent output
  variables.
\newblock {\em JMLR}, 6(Sep):1453--1484, 2005.

\bibitem{tsoumakas2006multi}
Grigorios Tsoumakas and Ioannis Katakis.
\newblock Multi-label classification: An overview.
\newblock {\em International Journal of Data Warehousing and Mining}, 3(3),
  2006.

\bibitem{vaswani2017attention}
Ashish Vaswani, Noam Shazeer, Niki Parmar, Jakob Uszkoreit, Llion Jones,
  Aidan~N Gomez, {\L}ukasz Kaiser, and Illia Polosukhin.
\newblock Attention is all you need.
\newblock In {\em Advances in Neural Information Processing Systems}, pages
  6000--6010, 2017.

\bibitem{wang2016cnn}
Jiang Wang, Yi Yang, Junhua Mao, Zhiheng Huang, Chang Huang, and Wei Xu.
\newblock Cnn-rnn: A unified framework for multi-label image classification.
\newblock In {\em Proceedings of the IEEE Conference on Computer Vision and
  Pattern Recognition}, pages 2285--2294, 2016.

\bibitem{feedbackprop_CVPR_2018}
Tianlu Wang, Kota Yamaguchi, and Vicente Ordonez.
\newblock Feedback-prop: Convolutional neural network inference under partial
  evidence.
\newblock In {\em IEEE Conference on Computer Vision and Pattern Recognition
  (CVPR)}, June 2018.

\bibitem{wang2017multi}
Zhouxia Wang, Tianshui Chen, Guanbin Li, Ruijia Xu, and Liang Lin.
\newblock Multi-label image recognition by recurrently discovering attentional
  regions.
\newblock In {\em Proceedings of the IEEE international conference on computer
  vision}, pages 464--472, 2017.

\bibitem{Wang_2017_ICCV}
Zhouxia Wang, Tianshui Chen, Guanbin Li, Ruijia Xu, and Liang Lin.
\newblock Multi-label image recognition by recurrently discovering attentional
  regions.
\newblock In {\em The IEEE International Conference on Computer Vision (ICCV)},
  Oct 2017.

\bibitem{hcp}
Y. {Wei}, W. {Xia}, M. {Lin}, J. {Huang}, B. {Ni}, J. {Dong}, Y. {Zhao}, and S.
  {Yan}.
\newblock Hcp: A flexible cnn framework for multi-label image classification.
\newblock {\em IEEE Transactions on Pattern Analysis and Machine Intelligence},
  38(9):1901--1907, Sep. 2016.

\bibitem{welinder2010caltech}
Peter Welinder, Steve Branson, Takeshi Mita, Catherine Wah, Florian Schroff,
  Serge Belongie, and Pietro Perona.
\newblock Caltech-ucsd birds 200.
\newblock 2010.

\bibitem{xie2018partial}
Ming-Kun Xie and Sheng-Jun Huang.
\newblock Partial multi-label learning.
\newblock In {\em Thirty-Second AAAI Conference on Artificial Intelligence},
  2018.

\bibitem{10.1109/ICCV.2011.6126300}
Xiangyang Xue, Wei Zhang, Jie Zhang, Bin Wu, Jianping Fan, and Yao Lu.
\newblock Correlative multi-label multi-instance image annotation.
\newblock In {\em Proceedings of the 2011 International Conference on Computer
  Vision}, ICCV '11, pages 651--658, USA, 2011. IEEE Computer Society.

\bibitem{bbox}
Hao Yang, {Joey Tianyi} Zhou, Yu Zhang, Bin-Bin Gao, Jianxin Wu, and Jianfei
  Cai.
\newblock Exploit bounding box annotations for multi-label object recognition.
\newblock In {Lourdes } Agapito, {Tamara } Berg, Jana Kosecka, and {Lihi }
  Zelnik-Manor, editors, {\em Proceedings - 29th IEEE Conference on Computer
  Vision and Pattern Recognition, CVPR 2016}, pages 280--288, United States of
  America, 2016. IEEE, Institute of Electrical and Electronics Engineers.

\bibitem{yeh2017learning}
Chih-Kuan Yeh, Wei-Chieh Wu, Wei-Jen Ko, and Yu-Chiang~Frank Wang.
\newblock Learning deep latent space for multi-label classification.
\newblock In {\em AAAI}, pages 2838--2844, 2017.

\bibitem{SRN}
Feng Zhu, Hongsheng Li, Wanli Ouyang, Nenghai Yu, and Xiaogang Wang.
\newblock Learning spatial regularization with image-level supervisions for
  multi-label image classification.
\newblock {\em 2017 IEEE Conference on Computer Vision and Pattern Recognition
  (CVPR)}, pages 2027--2036, 2017.

\end{thebibliography}
